\documentclass[lettersize,journal]{IEEEtran}
\usepackage{amsmath,amsfonts}
\usepackage{algorithm}
\usepackage{algpseudocode}
\usepackage{array}
\usepackage[caption=false,font=normalsize,labelfont=sf,textfont=sf]{subfig}
\usepackage{textcomp}
\usepackage{stfloats}
\usepackage{url}
\usepackage{verbatim}
\usepackage{graphicx}
\usepackage{cite}
\usepackage{amsmath}
\usepackage{amssymb}
\usepackage{xcolor}
\usepackage{booktabs}
\usepackage{multirow}
\usepackage{orcidlink}
\usepackage[capitalise]{cleveref}
\begin{document}

\title{A Generic Layer Pruning Method for Signal Modulation Recognition Deep Learning Models}

\author{Yao Lu\orcidlink{0000-0003-0655-7814}, Yutao Zhu\orcidlink{0009-0005-7154-1917}, Yuqi Li\orcidlink{0009-0002-7314-6863}, Dongwei Xu\orcidlink{0000-0003-2693-922X},~\IEEEmembership{Member, IEEE}, Yun Lin\orcidlink{0000-0003-1379-9301},~\IEEEmembership{Member, IEEE}, Qi Xuan\orcidlink{0000-0002-6320-7012},~\IEEEmembership{Senior Member,~IEEE}, Xiaoniu Yang\orcidlink{0000-0003-3117-2211}
        % <-this % stops a space
\thanks{This work was partially supported by the Key R\&D Program of Zhejiang under Grant 2022C01018 and by the National Natural Science Foundation of China under Grant U21B2001 and Grant 61973273. (Corresponding author: Qi Xuan)}% <-this % stops a space
\thanks{Yao Lu, Yutao Zhu and Dongwei Xu are with the Institute of Cyberspace Security, College of Information Engineering, Zhejiang University of Technology, Hangzhou, China, also with the Binjiang Institute of Artificial Intelligence, Zhejiang University of Technology, Hangzhou 310056, China (e-mail: yaolu.zjut@gmail.com, zhuyutao629@gmail.com, dongweixu@zjut.edu.cn).}
\thanks{Yuqi Li is currently engaged as a research intern at the Institute of Computing Technology, Chinese Academy of Sciences, Beijing, China (e-mail: yuqili010602@gmail.com).}
\thanks{Yun Lin is with the College of Information and Communication Engineering, Harbin Engineering University, Harbin, China (e-mail: linyun@hrbeu.edu.cn).}
\thanks{Qi Xuan is with the Institute of Cyberspace Security, College of Information Engineering, Zhejiang University of Technology, Hangzhou, China, also with the PCL Research Center of Networks and Communications, Peng Cheng Laboratory, Shenzhen 518000, China, and also with Utron Technology Company Ltd. (as Hangzhou Qianjiang Distinguished Expert), Hangzhou 310056, China (e-mail: xuanqi@zjut.edu.cn).}
\thanks{Xiaoniu Yang is with the Science and Technology on Communication Information Security Control Laboratory, Jiaxing, China (e-mail: yxn2117@126.com).}}

% The paper headers
\markboth{Journal of \LaTeX\ Class Files,~Vol.~14, No.~8, August~2021}%
{Shell \MakeLowercase{\textit{et al.}}: A Sample Article Using IEEEtran.cls for IEEE Journals}

% \IEEEpubid{0000--0000/00\$00.00~\copyright~2021 IEEE}
% Remember, if you use this you must call \IEEEpubidadjcol in the second
% column for its text to clear the IEEEpubid mark.

\maketitle

\begin{abstract}
With the successful application of deep learning in communications systems, deep neural networks are becoming the preferred method for signal classification. Although these models yield impressive results, they often come with high computational complexity and large model sizes, which hinders their practical deployment in communication systems. To address this challenge, we propose a novel layer pruning method. Specifically, we decompose the model into several consecutive blocks, each containing consecutive layers with similar semantics. Then, we identify layers that need to be preserved within each block based on their contribution. Finally, we reassemble the pruned blocks and fine-tune the compact model. Extensive experiments on five datasets demonstrate the efficiency and effectiveness of our method over a variety of state-of-the-art baselines, including layer pruning and channel pruning methods.
\end{abstract}


\section{Introduction}

\IEEEPARstart{A}{utomatic} Modulation Recognition (AMR) is an important research branch in the field of communications systems. Classical methods usually extract important hand-crafted features such as frequency, phase, constellation diagrams, high-order moments, and time-frequency diagrams, followed by applying machine learning methods for feature classification~\cite{walenczykowska2016type,li2019wavelet,triantafyllakis2017phasma,vuvcic2017cyclic,abdelmutalab2016automatic}. However, classical methods require extensive expertise and may lack high classification accuracy, which significantly limits their application, especially in complex scenarios.

Recent breakthroughs in deep learning have laid the foundation for developing high-performance models for AMR. For example, O'Shea et al.~\cite{o2018over} design a 1D-CNN model suitable for short radio signal classification based on the principles of the VGG~\cite{simonyan2014very} architecture. O'Shea et al.~\cite{o2016convolutional} develop a narrow 2D-CNN for radio modulation recognition by simultaneously considering the in-phase (I) and quadrature (Q) signal sequences in the time domain. Recently, Chen et al.~\cite{chen2021signet} introduce a sliding square operator called S2M, which automatically converts input signals into square feature matrices, facilitating the use of more complex models for signal classification. Although these models have achieved impressive results in signal classification tasks, they are often accompanied by high computational complexity and large model sizes. This results in slow inference speeds and makes them unsuitable for scenarios with limited computing resources, hindering their practical deployment in communication systems. 

To solve this problem, we need to compress the model while minimizing the loss in classification performance. Model pruning, as a model compression technique, provides a feasible solution for the above requirements. It can be roughly divided into weight pruning~\cite{zhang2018systematic,ma2020pconv}, channel pruning~\cite{meng2020pruning,chen2023rgp} and layer pruning~\cite{lu2022understanding,tang2023sr}. Given that weight pruning requires specialized hardware support and channel pruning is constrained by the depth of the original model, we concentrate on layer pruning.

In this paper, we propose a new layer pruning method. Specifically, we first calculate the similarity between any two layers of the pre-trained model to obtain the similarity matrix, then sum it row by row. Next, we apply Fisher's optimal segmentation to partition the model along its depth into multiple blocks based on the resulting vector. To identify important layers within each block that contribute significantly to the model's performance, we maintain the other blocks unchanged and enumerate all possible layer combinations within the selected block. For each layer combination, to maintain structural integrity, we retain the pre-trained parameters of the selected layer while randomly initializing the parameters of all other layers. Then, we utilize a training-free performance estimation method named SynFlow~\cite{tanaka2020pruning} to assess the contribution. Then, based on their contributions, we identify layers that need to be preserved within each block. Finally, we reassemble the pruned blocks and fine-tune the shallow model.

We conduct extensive experiments on five benchmarks, RML2016.10a~\cite{o2016radio}, RML2016.10a-high, Sig2019-12~\cite{chen2021signet}, Sig2019-12-high and RML2018.01a~\cite{o2018over}. The results demonstrate the superior performance of our method over existing channel pruning methods such as RFP~\cite{shao2021filter}, FPGM~\cite{he2019filter}, L1-norm~\cite{li2016pruning}, SFP~\cite{he2018soft} and BNP~\cite{chen2023channel}, as well as layer pruning methods random layer pruning, LCP-based pruning methods~\cite{chen2018shallowing,wang2019dbp} and SR-init~\cite{tang2023sr}.

% The idea of blocking is inspired by the different functions of different layers of the model.

To summarize, our main contributions are three-fold:
\begin{itemize}
\item[$\bullet$] We introduce a novel layer pruning method to slim deep learning models for AMR. Specifically, we first calculate the similarity matrix between layers, and then divide the pre-trained model into blocks based on the obtained similarity matrix by minimizing intra-block differences while maximizing inter-block differences. % , which consists of model partition, layer selection and model reassembly.
\item[$\bullet$] Subsequently, we identify layers that need to be preserved within each block based on their contribution. Finally, we reassemble the pruned blocks and fine-tune the compact model. To avoid the tedious training-validation process, we utilize a training-free performance estimation method named SynFlow to assess the contribution.
\item[$\bullet$] Extensive experiments demonstrate the efficiency and effectiveness of our method over a variety of state-of-the-art baselines on five datasets.
\end{itemize}

In the remainder of this paper, we first introduce related works on representational similarity and model pruning in \cref{sec:Related Works}. Preliminaries are presented in \cref{sec:Preliminaries}. Our method is detailed in \cref{sec:model purify}, and all relevant experiments are discussed in \cref{sec:Experiments}. Finally, the paper concludes in \cref{sec:Conclusion}.

\section{Related Works}
\label{sec:Related Works}
\textbf{Representational similarity} aims to calculate the similarity or difference between two internal representations of deep neural networks. By analyzing these representations, researchers can gain insights into how neural networks process and transform input data through their multiple layers. For example, Hardoon et al.~\cite{hardoon2004canonical} propose canonical correlation analysis (CCA) to learn a semantic representation to web images and their associated text. However, CCA is sensitive to perturbation when the condition number of $X$ or $Y$ is large. To this end, Raghu et al.~\cite{raghu2017svcca} and Morcos et al.~\cite{morcos2018insights} propose Singular Vector CCA and Projection Weighted CCA to reduce the sensitivity of CCA to perturbation, respectively. Later, Kornblith et al.~\cite{kornblith2019similarity} point out that these metrics can not measure meaningful similarities between representations of higher dimensions than the number of data points and propose a similarity index, namely centered kernel alignment (CKA), that does not suffer from this limitation. Recently, Chen et al.~\cite{chen2021graph} revisit the CKA and find that it implicitly constructs graphs to obtain similarity. Inspired by this, they introduce graph-based similarity, which explicitly constructs graphs based on the feature relationship between input samples to measure the similarity of deep learning models.

\textbf{Model pruning} has been widely acknowledged as a model compression technique to achieve acceleration on various platforms. It can be roughly divided into weight pruning~\cite{zhang2018systematic,ma2020pconv,aghli2021combining}, channel pruning~\cite{meng2020pruning,molchanov2019importance,he2017channel,zhuang2018discrimination} and layer pruning~\cite{lu2022understanding,tang2023sr,chen2018shallowing,huang2018data} based on the level of fine-grainedness. Specifically, weight pruning sets particular weights in a deep learning model to zero. Despite its high compression ratio, this method has limitations and is applicable only to specialized software~\cite{park2016faster} or hardware~\cite{han2016eie} devices. In contrast, channel pruning achieves speedup by discarding some filters from each layer of the model. Since filter pruning operates at the filter level, its capacity to prune parameters is constrained by the original model’s depth. Layer pruning, however, deletes entire layers, which is not constrained by this. Consequently, this paper primarily focuses on layer pruning.

The core of layer pruning lies in the selection of layers for removal, targeting those that contribute minimally to performance, without notably sacrificing overall performance. To this end, previous studies have developed various metrics to assess the importance of each layer. For example, Elkerdawy et al.~\cite{elkerdawy2020filter} use existing filter criteria to calculate a per-layer importance score in one-shot, subsequently pruning the least important layers and fine-tuning the shallower model. Chen et al.~\cite{chen2018shallowing} train Linear Classifier Probes (LCPs) to analyze the specific role of each convolutional layer in enhancing performance. Then they prune layers that contribute minimally to performance enhancement and subsequently retrain the pruned model with a knowledge distillation technique. However, the costs of training LCPs are quite high, especially for large datasets like ImageNet~\cite{deng2009imagenet}. To address this, Lu et al.~\cite{lu2022understanding} introduce modularity as a metric that requires no training for quantifying the class separation of intermediate representations. They then prune layers exhibiting negative or zero modularity growth. Elkerdawy et al.~\cite{elkerdawy2020one} construct proxy classifiers for each layer in a single pass on training dataset using imprinting, and then prune layers that exhibit the smallest accuracy difference compared to their preceding layer. Unlike previous studies that prune layers based on their importance, we assess the redundancy of layers by examining the similarity between them.

Aside from layer pruning, some effective layer compression methods also exist, which offer additional strategies for reducing model complexity. For example, Dror et al.~\cite{dror2021layer} propose Layer Folding to identify which activations can be removed with minimal impact on accuracy. This enables the merging of adjacent linear layers, thereby transforming deep models into shallower ones. Wu et al.~\cite{wu2023efficient} introduce De-Conv and Rem-ReLU to decouple convolutional layers and non-linear activation layers, making the model mergeable. Leveraging the property of linear mergeability of convolutional layers, they then losslessly merge the decoupled layers, achieving efficient layer compression. These methods are orthogonal to ours, allowing them to complement each other.

\section{Preliminaries}
\label{sec:Preliminaries}
% \textcolor{blue}{In this section, we first formulate layer pruning.}
\subsection{Mathematical Expression of Layer Pruning}
Before delving into our method, we first establish a formal mathematical formulation for layer pruning.

Suppose we have a pre-trained model $\mathcal{M}=L_{1} \circ L_{2} \cdots \circ L_{l}$ designed for task $T$, where $L_i$ denotes the $i$-th layer of $\mathcal{M}$ and $\circ$ denotes function composition. Considering computational constraints, we need to prune some layers in $\mathcal{M}$ while maximizing the performance of the pruned model $\mathcal{M}^*$. We can formulate it as an optimization problem:
\begin{equation}
\begin{aligned}
\mathcal{M}^*&=\max _{\mathcal{M}_p} P_T(\mathcal{M}_p), \\  
\text {s.t.} \quad \mathcal{M}_p=L_{i} &\circ L_{j} \cdots \circ L_{k}, \quad \text {cost}(\mathcal{M}_p) \leq C.
\end{aligned}
\end{equation}
Since $\mathcal{M}_p$ is composed of the remaining layers of $\mathcal{M}$, the total number of layers of $\mathcal{M}_p$ should be less than that of $\mathcal{M}$, and $1 \leq i < j < \cdots < k \leq l$. For two consecutive layers with dimension mismatch in $\mathcal{M}_p$, we adjust the feature size of the subsequent layer to align with that of the preceding layer. Besides, $P_T(\mathcal{M}_p)$ denotes the performance of $\mathcal{M}_p$ on task $T$, and $\text {cost}(\mathcal{M}_p) \leq C$ ensures that the computational overhead of $\mathcal{M}_p$ can not surpass $C$.

% \begin{equation}
% \mathcal{M}^*=\max _{\mathcal{M}_p} P_T(\mathcal{M}_p), \quad \text {s.t.} \quad |\mathcal{M}_p|\leq C.
% \end{equation}

\section{Method}
\label{sec:model purify}
In this section, we delve into the proposed layer pruning method. We first calculate the similarity matrix of layers, and then partition the pre-trained model into blocks by minimizing intra-block differences while maximizing inter-block differences. Subsequently, we identify layers that need to be preserved within each block and prune the remainder. Finally, we reassemble the factorized blocks and fine-tune the compact model.

\subsection{Model Partition}
A model partition refers to dividing a model into distinct sub-nets. In this study, we partition the model $\mathcal{M}$ along its depth into $k$ blocks $\mathcal{B} = \{B_1, \cdots, B_k\}$ so that each block is a stack of some consecutive layers. In that case, what is the basis for division? In this paper, CKA~\cite{kornblith2019similarity}, a technique for measuring representation similarity, serves as our basis for division. It is worth noting that CKA is not the only applicable similarity metric. In \cref{sec: Ablation Study}, we demonstrate that our method supports a variety of similarity metrics. Through analyzing the feature representation produced by each layer, we can identify layers that produce similar outputs and group them into the same block.
% $B_i = L_{i} \circ L_{i+1} \cdots \circ L_{i+n}$. Inspired by the hierarchical nature of deep neural networks~\cite{zeiler2014visualizing,wang2018visualizing}, we aim to partition the model according to their function level. For instance, low-level curves, high-level semantics. To differentiate level, we use similarity instead and thus introduce centered kernel alignment (CKA) to measure the similarity between a given pair of layers.

Specially, given a batch of examples $X_{b}$, the feature representations of $L_{i}$ and $L_{j}$ can be formulated as
\begin{equation}
\begin{aligned}
F_{i} &= L_{1} \circ L_{2} \cdots \circ L_{i}(X_{b}) \\
F_{j} &= L_{1} \circ L_{2} \cdots \circ L_{j}(X_{b}).
\end{aligned}
\end{equation}
Here, $F_i \in \mathbb{R} ^{b\times c_i \times h_i \times w_i}$ and $F_j \in \mathbb{R} ^{b\times c_j \times h_j \times w_j}$, where $b$ denotes the number of samples, while $c_i$, $h_i$ and $w_i$ represent the number of channels, height and width of layer $L_{i}$, respectively. To simplify the calculations, we apply a $\operatorname{flatten}$ operation to $F_i$ and $F_j$, resulting in $\hat{F_i}$ and $\hat{F_j}$.
\begin{equation}
    \hat{F} = \operatorname{flatten}(F),
\end{equation}
where the $\operatorname{flatten}(\cdot)$ operation transforms the tensor from $\mathbb{R} ^{b\times c \times h \times w}$ to $\mathbb{R} ^{b\times (c \times h \times w)}$, preserving the number of samples $b$ and combining the dimensions $c$, $h$, and $w$ into a single dimension. Subsequently, we calculate the gram matrices $K = \hat{F}_i \hat{F}_i^\mathsf{T}$ and $L = \hat{F}_j \hat{F}_j^\mathsf{T}$ based on $\hat{F}_{i}$ and $\hat{F}_{j}$. Next, we use Hilbert-Schmidt Independence Criterion~\cite{gretton2005measuring} ($\operatorname{HSIC}$) to calculate the statistical independence between $\hat{F}_{i}$ and $\hat{F}_{j}$, 
\begin{equation}
\operatorname{HSIC_{0}}(\hat{F}_{i}, \hat{F}_{j})=\frac{1}{(b-1)^2} \operatorname{tr}(\hat{F}_{i} H \hat{F}_{j} H)
\end{equation}
where $H=I_b-\frac{1}{b} \mathbf{1 1}^T$ is the centering matrix. However, $\operatorname{HSIC_{0}}$ exhibits an $O\left(\frac{1}{b}\right)$ bias~\cite{gretton2005measuring}, making it dependent on batch size $b$. To mitigate the influence of batch size, we replace $\operatorname{HSIC_{0}}$ with an unbiased one~\cite{song2012feature}:
\begin{equation}
\label{eq3}
\begin{aligned}
\operatorname{HSIC_{1}}(K, L)&=\\ \frac{1}{b(b-3)} &\left[\operatorname{tr}\left(\tilde{K} \tilde{L}\right)+\frac{\mathbf{1}^{\top} \tilde{K} \mathbf{1 1}^{\top} \tilde{L} \mathbf{1}}{(b-1)(b-2)}-\frac{2}{b-2} \mathbf{1}^{\top} \tilde{K} \tilde{L} \mathbf{1}\right],
\end{aligned}
\end{equation}
where $\tilde{K}=K\left(\mathbf{1 1^\mathsf{T}}-I_{b}\right)$ and $\tilde{L}=L\left(\mathbf{1 1^\mathsf{T}}-I_{b}\right)$. It is worth noting that $\operatorname{HSIC}$ is not invariant to isotropic scaling, but it can be made invariant through normalization. Therefore, CKA can be calculated as follows: 
\begin{equation}
\label{eq5} 
\operatorname{CKA}(K, L) = \frac{\operatorname{HSIC_{1}}(K, L)}{\sqrt{\operatorname{HSIC_{1}}(K, K)\operatorname{HSIC_{1}}(L, L)}}.
\end{equation}
Next, we compute the similarity between any two layers using \cref{eq5} iteratively, ultimately yielding a similarity matrix $\mathcal{S}$. Then we sum the matrix $\mathcal{S}$ row by row:  
\begin{equation}
\label{eq6} 
z_i = \sum_{j=1}^l \mathcal{S}_{ij}
\end{equation}
The resulting vector $\mathcal{Z}=\{z_1, \cdots, z_l\}$ contains the aggregated similarity of each layer, representing the degree of similarity between each layer and all other layers. Having obtained $\mathcal{Z}$, we utilize Fisher's optimal segmentation, a method for segmenting ordered sequences with the goal of minimizing the differences within segments and maximizing the differences between segments, to partition the model $\mathcal{M}$ along its depth into $k$ blocks. The overall process of model partition is illustrated in \cref{algorithm:Partition}.

\begin{algorithm}[t]
    \caption{Model Partition}
    \label{algorithm:Partition}
    \textbf{Input}: A batch of examples $X_{b}$, a pre-trained model $\mathcal{M}=L_{1} \circ L_{2} \cdots \circ L_{l}$ and number of blocks $k$.\\
    \textbf{Output}: A segmented pre-trained model $\mathcal{M}'$.\\
    \begin{algorithmic}[1] %[1] enables line numbers
    % \State \# Calculating the similarity matrix between layers
    \State Initializing $\mathcal{Z}=\{\}$
    \For{$i = 1$ to $l$}
        \State $F_{i} = L_{1} \circ L_{2} \cdots \circ L_{i}(X_{b})$
        \State $\hat{F}_i = \operatorname{flatten}(F_i)$
        \State $K = \hat{F}_i \hat{F}_i^T$
        \For{$j = 1$ to $l$}
            \State $F_{j} = L_{1} \circ L_{2} \cdots \circ L_{j}(X_{b})$
             \State $\hat{F}_j = \operatorname{flatten}(F_j)$
            \State $L = \hat{F}_j \hat{F}_j^T$
            \State \text{Using \cref{eq5} with $K$ and $L$ to calculate $\mathcal{S}_{ij}$.}
        \EndFor
    \EndFor
    \For{$i = 1$ to $l$}
        \State $z_i = \sum_{j=1}^l \mathcal{S}_{ij}$
        \State Putting $z_i$ into $\mathcal{Z}$
    \EndFor
    \State Using Fisher's optimal segmentation to partition $\mathcal{Z}$ into $k$ blocks.
    \end{algorithmic}
\end{algorithm}

\begin{algorithm}[t]
    \caption{Layer Selection and Model Reassembly}
    \label{algorithm:Layer Selection and Model Reassembly}
    \textbf{Input}: All possible layer combinations $\{\mathcal{C}_1, \cdots, \mathcal{C}_k\}$, a segmented pre-trained model $\mathcal{M}'$ and number of blocks $k$.\\
    \textbf{Output}: Optimal layer combination $\mathcal{M}_o$. \\
    \begin{algorithmic}[1] %[1] enables line numbers
    % \State \# Calculating the similarity matrix between layers
    \State Initializing $\mathcal{M}_o=\{\}$
    \For{$i = 1$ to $k$}
    % \State Initializing $B_i^o=\{\}$
    \State Initializing $\text{Best Acc}_i=0$
    \For{$j = 1$ to $|\mathcal{C}_i|$}
    \State The selected layer combination = $\mathcal{C}_i^j$
    \State \text{Retaining pre-trained parameters of selected layers.}
    \State \text{Randomly initializing parameters of other layers.}
    \State \text{Evaluating the accuracy $\text{Acc}_i^j$ using SynFlow.}
    \State \text{\textbf{if}} $\text{Acc}_i^j \geq \text{Best Acc}_i$ \text{\textbf{then}}
    \State \quad $\text{Best Acc}_i = \text{Acc}_i^j$
    \State \quad $B_i^o=\mathcal{C}_i^j$
    \State \text{\textbf{end if}}
    \EndFor
    \State Putting $B_i^o$ into $\mathcal{M}_o$
    \EndFor
    \end{algorithmic}
\end{algorithm}
\subsection{Layer Selection and Model Reassembly}
As we have partitioned the model into $k$ blocks, our next goal is to identify the layers that should be retained within each block.

Suppose $\mathcal{M}'=(L_{1} \circ \cdots L_{i}) \circ (L_{i+1} \circ \cdots L_{j}) \circ (L_{j+1} \circ \cdots L_{l})$ is a segmented pre-trained model and layers in the same bracket are in the same block. Our goal is to identify and retain important layers within each block that contribute significantly to the performance of the model. To achieve this goal, a direct method is keeping the other blocks unchanged, enumerating all possible layer combinations within the selected block, and selecting the optimal combination. Specifically, for block $B_i \in \mathcal{B}$ in $\mathcal{M}'$, we maintain a set $\mathcal{C}_i$ that includes all possible layer combinations within this block, except the empty set. If a block contains $m$ layers, the set $\mathcal{C}_i$ would have $\sum_{i=1}^m C_{m}^i$ elements, where $C_{m}^i$ denotes the total number of combinations obtained by randomly sampling $i$ layers from the block without replacement. Then we evaluate the contribution of each combination in $\mathcal{C}_i$ to the model's overall performance and pick the best combination. However, such an evaluation faces two major limitations:

\textbf{(1)} Verifying the contribution of selected layers to the model requires reconstructing the corresponding blocks of the original model so that only these selected layers are included in the blocks. For example, suppose $L_1$ and $L_2$ are the selected layers in block $B_1$, then the model should be adjusted to $(L_{1} \circ L_{2}) \circ (L_{i+1} \circ \cdots L_{j}) \circ (L_{j+1} \circ \cdots L_{l})$. For each combination of layers, we need to reconstruct the block where those layers are located, which is very laborious.

\textbf{(2)} Besides, evaluating the contribution of each combination through training and validation processes is also impractical, especially for models with a large number of layers.

To address these issues, we propose the following solutions: For issue $1$, inspired by Tang et al.~\cite{tang2023sr}, we retain the pre-trained parameters of the selected layer while randomly initializing the parameters of all other layers to simulate evaluating the importance of the selected layer. This process can be formalized as:
\begin{equation}
\label{eq7}
% \max _{\mathcal{M}'} P_T(\mathcal{M}'), \\ 
\mathcal{M}'=(L_{1} \circ L_{2} \circ \cdots \hat{L_{i}}) \circ (\hat{L_{i+1}} \circ \cdots \hat{L_{j}}) \circ (\hat{L_{j+1}} \circ \cdots \hat{L_{l}}),
\end{equation}
where $\hat{L_{i}}$ denotes that $L_{i}$ is initialized using kaiming initialization~\cite{he2015delving}. Such operation avoids the need to make different adjustments to the model when selecting different layers. As for issue $2$, we substitute the expensive training and validation processes with a training-free performance estimation method SynFlow~\cite{tanaka2020pruning}, significantly reducing the cost of verifying the contribution of each combination.

% As for issue $2$, we decrease the number of combinations by increasing the number of blocks, thereby ensuring that each block contains fewer layers. In \cref{sec: Ablation Study}, we demonstrate that changing the number of blocks does not affect the effectiveness of our method.

Having addressed the aforementioned challenges, we elaborate on the process of layer selection: For each block in $\mathcal{M}'$, we maintain a set $\mathcal{C}_i$ that includes all possible layer combinations within this block, except the empty set. We first initialize an empty set $\mathcal{M}_o$ to store the optimal layer combination of the segmented pre-trained model $\mathcal{M}'$. Besides, we initialize the best accuracy $\text{Best Acc}_i=0$. Then for each layer combination $\mathcal{C}_i^j$ in $\mathcal{C}_i$, we retain the pre-trained parameters of the selected layer while randomly initializing the parameters of all other layers. Next, we use the training-free performance estimation method SynFlow to evaluate the accuracy $\text{Acc}_i^j$ of this model. Then for each block $B_i$, we keep the optimal layer combination $B_i^o$ with the best performance and put it into $\mathcal{M}_o$. Finally, we repeat the above operation for each block to obtain the optimal layer combination. The overall process is illustrated in \cref{algorithm:Layer Selection and Model Reassembly}.

Having identified the important layers within each block, we proceed to integrate these layers into a compact model. Since selection destroys the original layer structure, dimensional mismatch may occur during the model reassembly. If this happens, we adjust the subsequent layer to align with the dimension of the previous layer. Finally, we fine-tune the compact model.
% In order to facilitate understanding, we give the following examples. Suppose $F_i$ and $F_{i}$ are two selected layers in block $B_i$. 
% $\mathcal{M} = B_{1} \circ B_{2} \cdots \circ B_{k}$

% Table generated by Excel2LaTeX from sheet 'Sheet1'  resnet56，k=3,高信噪比；resnet110，k=7，vgg16，k=3，高信噪比									
\begin{table*}[t]
  \centering
  \caption{Pruning results of ResNet56, ResNet110 and VGG16 on three high SNR datasets. PR denotes the pruning rate.}
   \resizebox{0.99\textwidth}{!}{
    \begin{tabular}{c|ccccccccc}
    \toprule
    \multicolumn{1}{c}{Model} & Dataset & Original Acc(\%) & PR & Acc(\%) &  $\triangle$Acc(\%) &  $\triangle$Flops &  $\triangle$Params & Flops PR & Params PR \\
    \midrule
    \multirow{12}[6]{*}{ResNet56} & \multirow{4}[2]{*}{RML2016.10a-high} & \multirow{4}[2]{*}{90.32 } & 25\%  & 91.06  & 0.74 $\uparrow$  & -10.69M & -212.99K & 25.5\% & 25.0\% \\
          &       &       & 50\%  & 91.13  & 0.81 $\uparrow$  & -24.92M & -578.36K & 59.4\% & 67.8\% \\
          &       &       & 75\%  & 90.76  & 0.44 $\uparrow$ & -34.44M & -735.93K & 82.2\% & 86.3\% \\
          &       &       & 89\%  & 90.25  & -0.07 $\downarrow$ & -38.01M & -777.72K & 90.7\% & 91.2\% \\
\cmidrule{2-10}          & \multirow{4}[2]{*}{Sig2019-12-high} & \multirow{4}[2]{*}{97.42} & 25\%  & 97.57  & 0.15 $\uparrow$ & -52.13M & -351.62K & 31.1\% & 41.2\% \\
          &       &       & 50\%  & 97.18  & -0.24 $\downarrow$ & -94.83M & -592.38K & 56.6\% & 69.5\% \\
          &       &       & 75\%  & 97.26  & -0.16 $\downarrow$ & -128.32M & -625.09K & 76.5\% & 73.3\% \\
          &       &       & 86\%  & 96.94  & -0.48 $\downarrow$ & -152.04M & -777.73K & 90.7\% & 91.2\% \\
\cmidrule{2-10}          & \multirow{4}[2]{*}{RML2018.01a-high} & \multirow{4}[2]{*}{89.70 } & 25\%  & 89.57  & -0.13 $\downarrow$ & -94.83M & -296.19K & 28.3\% & 34.7\% \\
          &       &       & 50\%  & 88.10  & -1.60 $\downarrow$ & -161.54M & -384.45K & 48.2\% & 45.0\% \\
          &       &       & 75\%  & 86.64  & -3.06 $\downarrow$ & -266.01M & -694.40K & 79.3\% & 81.3\% \\
          &       &       & 89\%  & 84.07  & -5.63 $\downarrow$ & -304.09M & -777.73K & 90.7\% & 91.1\% \\
    \midrule
    \multirow{12}[6]{*}{ResNet110} & \multirow{4}[2]{*}{RML2016.10a-high} & \multirow{4}[2]{*}{90.37 } & 25\%  & 90.86  & 0.49 $\uparrow$ & -21.35M & -0.55M & 25.2\% & 31.8\% \\
          &       &       & 50\%  & 91.02  & 0.65 $\uparrow$ & -41.61M & -0.78M & 49.1\% & 45.1\% \\
          &       &       & 75\%  & 90.96  & 0.59 $\uparrow$ & -62.98M & -1.25M & 74.4\% & 72.3\% \\
          &       &       & 87\%  & 90.75  & 0.38 $\uparrow$ & -74.85M & -1.51M & 88.4\% & 87.3\% \\
\cmidrule{2-10}          & \multirow{4}[2]{*}{Sig2019-12-high} & \multirow{4}[2]{*}{97.09 } & 25\%  & 97.32  & 0.23 $\uparrow$ & -85.49M & -0.47M & 25.2\% & 27.2\% \\
          &       &       & 50\%  & 97.42  & 0.33 $\uparrow$ & -175.80M & -0.91M & 51.9\% & 52.6\% \\
          &       &       & 75\%  & 97.50  & 0.41 $\uparrow$ & -261.29M & -1.37M & 77.1\% & 79.2\% \\
          &       &       & 87\%  & 97.17  & 0.08 $\uparrow$ & -299.30M & -1.55M & 88.4\% & 89.6\% \\
\cmidrule{2-10}          & \multirow{4}[2]{*}{RML2018.01a-high} & \multirow{4}[2]{*}{86.73 } & 25\%  & 90.39  & 3.66 $\uparrow$ & -142.48M & -0.34M & 21.0\% & 19.7\% \\
          &       &       & 50\%  & 89.46  & 2.73 $\uparrow$ & -407.83M & -1.32M & 60.2\% & 76.3\% \\
          &       &       & 75\%  & 88.18  & 1.45 $\uparrow$ & -532.02M & -1.42M & 78.5\% & 82.1\% \\
          &       &       & 87\%  & 83.63  & -3.10 $\downarrow$ & -608.04M & -1.61M & 89.8\% & 93.1\% \\
    \midrule
    \multirow{3}[6]{*}{VGG16} & RML2016.10a-high & 90.42  & 56\%  & 89.37  & -1.05 $\downarrow$ & -0.72G & -6.05M & 80.0\% & 68.4\% \\
\cmidrule{2-10}          & Sig2019-12-high & 97.20  & 33\%  & 97.08  & -0.12 $\downarrow$ & -1.42G & -3.54M & 43.6\% & 29.5\% \\
\cmidrule{2-10}          & RML2018.01a-high & 83.65  & 44\%  & 82.58  & -1.07 $\downarrow$ & -3.71G & -5.91M & 58.1\% & 36.5\% \\
    \bottomrule
    \end{tabular}}
  \label{highsnr}%
\end{table*}%

% \textbf{Verifying the training-free proxy}. As the first attempt to apply the NASWOT to measure model transfer-ability, we verify its efficacy before applying it to DeRy task. We adopt the score to rank 10 pre-trained models on 8 image classification tasks.

\section{Experiments}
\label{sec:Experiments}
In this section, we verify the effectiveness of the proposed method on various datasets.

\subsection{Dataset}
To evaluate the effectiveness of our method, we conduct experiments on five signal modulation classification datasets RML2016.10a~\cite{o2016radio}, RML2016.10a-high, Sig2019-12~\cite{chen2021signet}, Sig2019-12-high and RML2018.01a~\cite{o2018over}.

\textbf{RML2016.10a} is a synthetic dataset generated using GNU Radio~\cite{blossom2004gnu}, containing 11 modulation types and a total of 220,000 signals ($1000$ samples per modulation type of each signal-to-noise ratio (SNR)). Specifically, BPSK, QPSK, 8PSK, 16QAM, 64QAM, BFSK, CPFSK and PAM4 for digital modulations, and WB-FM, AM-SSB, and AM-DSB for analog modulations. The SNRs span from $-20$dB to $18$dB in $2$dB intervals, with each signal having a length of 128. We partition the dataset into a training set, validation set, and test set in a ratio of 6:2:2.

\textbf{RML2016.10a-high} is a subset of RML2016.10a, consisting of samples with SNRs ranging from $10$dB to $18$dB. Compared to RML2016.10a, RML2016.10a-high is easier to classify~\cite{chen2021signet}.

\textbf{Sig2019-12} contains longer signals simulated by Chen et al.~\cite{chen2021signet}. It consists of 12 modulation types, including BPSK, QPSK, 8PSK, OQPSK, 2FSK, 4FSK, 8FSK, 16QAM, 32QAM, 64QAM, 4PAM, and 8PAM. The SNRs span from $-20$dB to $30$dB in $2$dB intervals, with each signal having a length of 512. The total size of Sig2019-12 is $468000$ ($1500$ samples per modulation type of each SNR).  We also partition the dataset into a training set, validation set, and test set in a ratio of 6:2:2.

\textbf{Sig2019-12-high} is a subset of Sig2019-12, consisting of samples with SNRs ranging from $10$dB to $30$dB. This subset is easier to classify compared to the full SNR range~\cite{chen2021signet}.

\textbf{RML2018.01a-high} RML2018.01a simulates a wireless channel and is collected from a laboratory environment. It consists of 24 modulation types, including OOK, 4ASK, 8ASK, BPSK, QPSK, 8PSK, 16PSK, 32PSK, 16APSK, 32APSK, 64APSK, 128APSK, 16QAM, 32QAM, 64QAM, 128QAM, 256QAM, AM-SSB-WC, AM-SSB-SC, AM-DSB-WC, AM-DSB-SC, FM, GMSK, OQPSK. The SNRs span from $-20$dB to $30$dB in $2$dB intervals, with each signal having a length of 1024. The total size of the dataset is $2555904$. Due to its large size, we use $10$dB to $30$dB SNR samples for experiments ($1056000$ samples in total). We partition the dataset into training and test sets in a ratio of 8:2.

% \textbf{CIFAR10} consists of 60,000 $32\times32$ RGB images, divided into 50,000 for training and 10,000 for testing. It contains images from 10 classes: airplane, automobile, bird, cat, deer, dog, frog, horse, ship, and truck.

% Table generated by Excel2LaTeX from sheet 'Sheet1' resnet56，k=3,全信噪比；resnet110，k=7，全信噪比									
\begin{table*}[t]
  \centering
  \caption{Pruning results of ResNet56, ResNet110 and VGG16 on three full SNR datasets. PR denotes the pruning rate.}
   \resizebox{0.99\textwidth}{!}{
    \begin{tabular}{c|ccccccccc}
    \toprule
    \multicolumn{1}{c}{Model} & Dataset & Original Acc(\%) & PR & Acc(\%) & $\triangle$Acc(\%) & $\triangle$Flops & $\triangle$Params & Flops PR & Params PR \\
    \midrule
    \multirow{8}[4]{*}{ResNet56} & \multirow{4}[2]{*}{RML2016.10a} & \multirow{4}[2]{*}{62.06 } & 25\%  & 62.37  & 0.31 $\uparrow$ & -10.70M & -199.10K & 25.6\% & 23.3\% \\
          &       &       & 50\%  & 61.80  & -0.26 $\downarrow$ & -21.37M & -439.87K & 51.0\% & 51.6\% \\
          &       &       & 75\%  & 61.32  & -0.74 $\downarrow$ & -32.06M & -675.83K & 76.5\% & 79.2\% \\
          &       &       & 86\%  & 60.90  & -1.16 $\downarrow$ & -36.82M & -759.16K & 87.8\% & 89.0\% \\
\cmidrule{2-10}          & \multirow{4}[2]{*}{Sig2019-12} & \multirow{4}[2]{*}{67.91 } & 25\%  & 68.12  & 0.21 $\uparrow$ & -47.48M & -268.42K & 28.3\% & 31.5\% \\
          &       &       & 50\%  & 67.60  & -0.31 $\downarrow$ & -80.77M & -384.45K & 48.2\% & 45.1\% \\
          &       &       & 75\%  & 67.19  & -0.72 $\downarrow$ & -132.77M & -694.40K & 79.2\% & 81.4\% \\
          &       &       & 89\%  & 64.93  & -2.98 $\downarrow$ & -152.04M & -777.73K & 90.7\% & 91.2\% \\
    \midrule
    \multirow{8}[4]{*}{ResNet110} & \multirow{4}[2]{*}{RML2016.10a} & \multirow{4}[2]{*}{61.97 } & 25\%  & 62.36  & 0.39 $\uparrow$ & -20.21M & -0.38M & 23.9\% & 22.0\% \\
          &       &       & 50\%  & 61.85  & -0.12 $\downarrow$ & -47.44M & -1.16M & 56.0\% & 67.1\% \\
          &       &       & 75\%  & 61.89  & -0.08 $\downarrow$ & -65.33M & -1.37M & 76.0\% & 79.2\% \\
          &       &       & 87\%  & 61.73  & -0.24 $\downarrow$ & -74.83M & -1.56M & 88.4\% & 90.2\% \\
\cmidrule{2-10}          & \multirow{4}[2]{*}{Sig2019-12} & \multirow{4}[2]{*}{65.67 } & 25\%  & 66.63  & 0.96 $\uparrow$ & -85.49M & -0.47M & 25.2\% & 27.2\% \\
          &       &       & 50\%  & 66.66  & 0.99 $\uparrow$ & -142.97M & -0.42M & 42.2\% & 25.3\% \\
          &       &       & 75\%  & 66.31  & 0.64 $\uparrow$ & -251.95M & -1.22M & 74.4\% & 70.5\% \\
          &       &       & 87\%  & 65.97  & 0.30 $\uparrow$ & -299.40M & -1.49M & 88.4\% & 86.1\% \\
    \midrule
    \multirow{2}[4]{*}{VGG16} & RML2016.10a & 58.70  & 67\%  & 57.91  & -0.79 $\downarrow$ & -0.81G & -6.08M & 90.0\% & 68.7\% \\
\cmidrule{2-10}          & Sig2019-12 & 65.32  & 67\%  & 64.43  & -0.89 $\downarrow$ & -2.93G & -6.09M & 89.9\% & 50.8\% \\
    \bottomrule
    \end{tabular}}
  \label{fullsnr}%
\end{table*}%
 
\subsection{Implementation details}
To verify the effectiveness of our proposed method, we use typical convolutional neural networks designed for the AMR task (ResNet56~\cite{he2016deep}, ResNet110~\cite{he2016deep} and VGG16~\cite{simonyan2014very}) as models for pruning and compare our proposed method with channel pruning methods RFP~\cite{shao2021filter}, FPGM~\cite{he2019filter}, L1-norm~\cite{li2016pruning}, SFP~\cite{he2018soft} and BNP~\cite{chen2023channel}, as well as layer pruning methods random layer pruning, LCP-based pruning methods~\cite{chen2018shallowing,wang2019dbp} and SR-init~\cite{tang2023sr}. We briefly summarize channel pruning baselines as follows:
\begin{itemize}
  \item[$\bullet$] RFP~\cite{shao2021filter} selects important filters based on the information entropy of feature maps.
  \item[$\bullet$] FPGM~\cite{he2019filter} prunes redundant filters based on the geometric median~\cite{fletcher2008robust} of the filters.
  \item[$\bullet$] L1-norm~\cite{li2016pruning} measures the relative importance of a filter in each layer by calculating its ${\ell _1}$-norm.
  \item[$\bullet$] SFP~\cite{he2018soft} dynamically prunes the filters using ${\ell _2}$-norm in a soft manner.
  \item[$\bullet$] BNP~\cite{chen2023channel} measures the channel importance using the $\gamma$ scale factor of the batch normalization layer.
\end{itemize}
Then we introduce layer pruning methods:
\begin{itemize}
  \item[$\bullet$] Random layer pruning serves as a baseline to verify the effectiveness of layer pruning.
  \item[$\bullet$] LCP-based pruning methods~\cite{chen2018shallowing,wang2019dbp} use linear classifier probes to delete unimportant layers and adapt different knowledge distillation techniques to recover the performance.
  \item[$\bullet$] SR-init~\cite{tang2023sr} prunes layers that are not sensitive to stochastic re-initialization.
\end{itemize}
Since these methods do not conduct experiments on signal modulation classification datasets, we reproduce these methods ourselves. For a fair comparison, we set the same hyperparameter settings to fine-tune pruned models. Specifically, we set the initial learning rate, batch size and epoch to 0.001, 128 and 50, respectively. The learning rate is decayed by a factor of 0.8 every $10$ epochs. As for our method, we sample $500$ instances to compute the representational similarity and the resultant value is determined by averaging across $5$ batches. Unless otherwise specified, we set $k=3$.

\subsection{Results and Analysis}
\textbf{Results on high SNR datasets.}
First, we conduct experiments on high SNR datasets, such as RML2016.10a-high, Sig2019-12-high and RML2018.01a-high. Compared to full SNR datasets, high SNR datasets are easier to classify~\cite{chen2021signet}. In this study, we focus on pruning ResNet56, ResNet110 and VGG16 under different pruning rates (25\%, 50\%, 75\% and maximum pruning rate). For VGG16, we validate only the maximum pruning rate to maintain structural integrity due to its layer limitations. Specifically, we set $k=3$ for ResNet56 and VGG16, and $k=7$ for ResNet110. The experimental results are presented in \cref{highsnr}. As for RML2016.10a-high and Sig2019-12-high, we find that even after pruning 75\% or more of the layers, the accuracy of ResNet56 and ResNet110 are barely affected and may even improve. For instance, pruning 75\% of layers in ResNet110 can achieve an improvement of 0.59\% on RML2016.10a-high and 0.41\% on Sig2019-12-high. This improvement indicates that these models do have redundancy, highlighting the need for pruning. As for the more complex RML2018.01a-high dataset, pruning 25\%, 50\% and 75\% of layers in ResNet110 can achieve an improvement of 3.66\%, 2.73\% and 1.45\%, respectively. These improvements can be attributed to the high complexity of the model relative to the dataset, which leads to overfitting~\cite{he2016deep}. In contrast, ResNet56 experiences an obvious decrease in accuracy at high pruning rates (1.60\% under the 50\% pruning rate, 3.06\% under the 75\% pruning rate, 5.63\% under the 89\% pruning rate). For VGG16, to maintain the integrity of the model structure, we ensure that at least one convolutional layer remains in each structural layer. This results in a lower pruning rate for VGG16. After reducing 33-55\% of the layers, the accuracy drop of VGG16 on the three datasets is still affordable (1.05\% on RML2016.10a-high, 0.12\% on Sig2019-12-high, and 1.07\% on RML2018.01a-high). In general, the abovementioned results indicate that our method can effectively eliminate redundant layers to simplify the model while maintaining the accuracy.

% Table generated by Excel2LaTeX from sheet 'Sheet1' 通道剪枝方法对比，数据集为全信噪比的RML2016.10a和Sig2019-12＋RML2018.01a，不做说明默认K=3										
\begin{table*}[htbp]
  \centering
  \caption{Comparison to channel pruning methods. PR denotes the pruning rate. \textbf{Bold entries} are best results.}
   \resizebox{0.99\textwidth}{!}{
    \begin{tabular}{c|cccccccccc}
    \toprule
    \multicolumn{1}{c}{Model} & Dataset & PR & Method & Original Acc(\%) & Acc(\%) & $\triangle$Acc(\%) & $\triangle$FLOPs & $\triangle$Params & FLOPs PR & Params PR \\
    \midrule
    \multirow{18}[5]{*}{ResNet56} & \multirow{6}[1]{*}{RML2016.10a} & \multirow{6}[1]{*}{86\%} & RFP   & 60.70  & 54.72  & -5.98 $\downarrow$ & -33.74M & -749.19K & 80.5\% & 87.9\% \\
          &       &       & FPGM  & 62.61  & 59.43  & -3.18 $\downarrow$ & -34.29M & -765.76K & 81.8\% & 89.8\% \\
          &       &       & L1-norm & 62.56  & 60.70  & -1.86 $\downarrow$ & -36.51M & -743.65K & 87.1\% & 87.2\% \\
          &       &       & SFP & 60.70  & 58.72  & -1.98 $\downarrow$ & -19.77M & -381.94K & 47.2\% & 44.8\% \\
          &       &       & BNP  & 62.63  & 61.39  & -1.24 $\downarrow$ & -36.51M & -743.65K & 87.1\% & 87.2\% \\
          &       &       & Ours & 62.06  & 61.10  & \textbf{-0.96} $\downarrow$ & -36.82M & -759.16K & 87.8\% & 89.0\% \\
\cmidrule{2-11}          & \multirow{6}[2]{*}{Sig2019-12} & \multirow{6}[2]{*}{89\%} & RFP   & 66.20  & 55.21  & -10.99 $\downarrow$ & -161.57M & -765.60K & 96.4\% & 89.8\% \\
          &       &       & FPGM  & 66.22  & 50.23  & -15.99 $\downarrow$ & -162.48M & -778.81K & 96.9\% & 91.3\% \\
          &       &       & L1-norm & 67.16  & 64.13  & -3.03 $\downarrow$ & -163.60M & -761.35K & 97.6\% & 89.3\% \\
          &       &       & SFP & 66.22  & 17.22  & -49.00 $\downarrow$ & -66.57M & -483.90K & 39.7\% & 56.7\% \\
          &       &       & BNP  & 67.16  & 61.53  & -5.63 $\downarrow$ & -163.60M & -761.35K & 97.6\% & 89.3\% \\
          &       &       & Ours  & 67.91  & 64.93  & \textbf{-2.98} $\downarrow$ & -152.04M & -777.73K & 90.7\% & 91.2\% \\
\cmidrule{2-11}          & \multirow{6}[2]{*}{RML2018.01a-high} & \multirow{6}[2]{*}{89\%} & RFP   & 92.00  & 42.72  & -49.28 $\downarrow$ & -328.70M & -768.86K & 98.0\% & 90.2\% \\
          &       &       & FPGM  & 91.95  & 79.56  & -12.39 $\downarrow$ & -329.87M & -778.03K & 98.4\% & 91.2\% \\
          &       &       & L1-norm & 92.56  & 77.95  & -14.61 $\downarrow$ & -331.28M & -760.57K & 98.8\% & 89.2\% \\
          &       &       & SFP & 91.95  & 32.75  & -59.20 $\downarrow$ & -204.92M & -483.12K & 61.1\% & 56.7\% \\
          &       &       & BNP  & 92.00  & 82.79  & -9.21 $\downarrow$ & -331.28M & -760.57K & 98.8\% & 89.2\% \\
          &       &       & Ours  & 89.70  & 84.07  & \textbf{-5.63} $\downarrow$ & -304.09M & -777.73K & 90.7\% & 91.1\% \\
    \midrule
    \multirow{18}[6]{*}{ResNet110} & \multirow{6}[2]{*}{RML2016.10a} & \multirow{6}[2]{*}{93\%} & RFP   & 62.50  & 29.73  & -32.77 $\downarrow$ & -74.78M & -1.63M & 88.3\% & 94.2\% \\
          &       &       & FPGM  & 62.47  & 60.11  & -2.36 $\downarrow$ & -77.24M & -1.63M & 91.2\% & 94.2\% \\
          &       &       & L1-norm & 62.47  & 53.92  & -8.55 $\downarrow$ & -79.08M & -1.62M & 93.4\% & 93.6\% \\
          &       &       & SFP & 62.47  & 28.62  & -26.85 $\downarrow$ & -46.66M & -1.02M & 55.1\% & 59.0\% \\
          &       &       & BNP  & 62.47  & 42.77  & -19.70 $\downarrow$ & -79.08M & -1.62M & 93.4\% & 93.6\% \\
          &       &       & Ours  & 61.97  & 61.00  & \textbf{-0.97} $\downarrow$ & -78.41M & -1.58M & 92.6\% & 91.3\% \\
\cmidrule{2-11}          & \multirow{6}[2]{*}{Sig2019-12} & \multirow{6}[2]{*}{93\%} & RFP   & 64.70  & 9.87  & -54.83 $\downarrow$ & -329.20M & -1.62M & 97.2\% & 93.6\% \\
          &       &       & FPGM  & 64.65  & 60.27  & -4.38 $\downarrow$ & -326.59M & -1.63M & 96.4\% & 94.2\% \\
          &       &       & L1-norm & 64.70  & 54.58  & -10.12 $\downarrow$ & -333.13M & -1.62M & 98.3\% & 93.6\% \\
          &       &       & SFP & 64.65  & 8.56  & -56.09 $\downarrow$ & -258.29M & -1.02M & 76.3\% & 59.0\% \\
          &       &       & BNP  & 64.70  & 58.09  & -6.61 $\downarrow$ & -333.13M & -1.62M & 98.3\% & 93.6\% \\
          &       &       & Ours  & 65.67  & 65.90  & \textbf{0.23} $\uparrow$ & -313.62M & -1.58M & 92.6\% & 91.3\% \\
\cmidrule{2-11}          & \multirow{6}[2]{*}{RML2018.01a-high} & \multirow{6}[2]{*}{93\%} & RFP   & 90.50  & 66.80  & -23.70 $\downarrow$ & -657.70M & -1.62M & 98.5\% & 93.6\% \\
          &       &       & FPGM  & 90.50  & 71.71  & -18.79 $\downarrow$ & -664.84M & -1.63M & 98.1\% & 94.2\% \\
          &       &       & L1-norm & 90.50  & 75.99  & -14.51 $\downarrow$ & -661.87M & -1.62M & 99.2\% & 93.6\% \\
          &       &       & SFP & 90.50  & 76.04  & -14.46 $\downarrow$ & -529.92M & -1.02M & 78.2\% & 59.0\% \\
          &       &       & BNP  & 90.50  & 80.85  & -9.65 $\downarrow$ & -661.87M & -1.62M & 99.2\% & 93.6\% \\
          &       &       & Ours  & 86.73  & 83.79  & \textbf{-2.94} $\downarrow$ & -646.19M & -1.66M & 95.4\% & 96.0\% \\
    \bottomrule
    \end{tabular}}
  \label{channelpruning}%
\end{table*}%

\textbf{Results on full SNR datasets.} Compared to high SNR datasets, full SNR datasets are more complex. Therefore, we conduct experiments on full SNR datasets, such as RML2016.10a and Sig2019-12. Similarly, we prune ResNet56, ResNet110 and VGG16 under different pruning rates (25\%, 50\%, 75\% and maximum pruning rate). As shown in \cref{fullsnr}, we find that the experimental results are almost the same as before, which demonstrates that when confronted with more complex datasets, our method is capable of pruning the model effectively while ensuring that the accuracy is maintained (-0.24\% under 87\% pruning rate on RML2016.10a and 0.30\% under 87\% pruning rate on Sig2019-12, etc.).

\textbf{Comparison to channel pruning methods.} In the previous paragraphs, we have demonstrated the effectiveness of our method on both high SNR and full SNR datasets. Here, we compare our method with channel pruning methods. Since some baselines do not include experiments with VGG16 pruning, we only use ResNet56 and ResNet110 for comparison. For ease of comparison, we uniformly use the highest pruning rate achievable by our method for the experiments. Except for ResNet56 on RML2016.10a, which uses $k=4$, all other models use $k=3$ by default. We compare five  baselines: RFP~\cite{shao2021filter}, FPGM~\cite{he2019filter}, L1-norm~\cite{li2016pruning}, SFP~\cite{he2018soft} and BNP~\cite{chen2023channel}. The experimental results are shown in \cref{channelpruning}. Specifically, our method results in a performance drop of only 0.96\% on RML2016.10a when pruning 87\% of the layers in ResNet56, while the best method (BNP) that prunes 87\% of the channels experiences a performance drop of 1.24\%. Besides, parameters and FLOPs reductions are also greater than those achieved by BNP. It is worth noting that when the compression rate is high, the model's performance drops sharply with some pruning methods. For instance, pruning ResNet56 on Sig2019-12 and RML2018.01a-high with SFP results in a performance drop of 49.00\% and 59.20\%, respectively. The same phenomenon occurs on pruning ResNet110 with RFP and SFP. We believe this is because some channel pruning methods cause the model structure to collapse by pruning too many channels at a high compression rate, which leads to a substantial decrease in the accuracy of the pruned model. In contrast, our layer pruning method does not have this phenomenon. Overall, our method outperforms these channel pruning baselines, demonstrating the advantages of our approach.

% Table generated by Excel2LaTeX from sheet 'Sheet1' 层剪枝方法对比，数据集为数据集为全信噪比的RML2016.10a和Sig2019-12＋RML2018.01a，不做说明默认K=3											
\begin{table*}[htbp]
  \centering
  \caption{Comparison to layer pruning methods. PR denotes the pruning rate. \textbf{Bold entries} are best results.}
  \resizebox{0.99\textwidth}{!}{
    \begin{tabular}{c|cccccccccc}
    \toprule
    \multicolumn{1}{c}{Model} & Dataset & PR & Method & Original Acc(\%) & Acc(\%) & $\triangle$Acc(\%) & $\triangle$FLOPs & $\triangle$Params & FLOPs PR & Params PR \\
    \midrule
    \multirow{12}[5]{*}{ResNet56} & \multirow{4}[2]{*}{RML2016.10a} & \multirow{4}[2]{*}{86\%} & Random & 62.06  & 60.08  & -1.98 $\downarrow$ & -35.64M & -703.74K & 85.0\% & 82.5\% \\
          &       &       & LCP   & 62.06  & 60.84  & -1.22 $\downarrow$ & -36.81M & -773.05K & 87.8\% & 90.6\% \\
          &       &       & SR-init & 62.06  & 61.02  & -1.04 $\downarrow$ & -35.64M & -703.74K & 85.0\% & 82.5\% \\
          &       &       & Ours & 62.06  & 61.10  & \textbf{-0.96} $\downarrow$ & -36.82M & -759.16K & 87.8\% & 89.0\% \\
\cmidrule{2-11}          & \multirow{4}[2]{*}{Sig2019-12} & \multirow{4}[2]{*}{89\%} & Random & 67.91  & 61.77  & -6.14 $\downarrow$ & -152.04M & -777.73K & 90.7\% & 91.2\% \\
          &       &       & LCP   & 67.91  & 64.33  & -3.58 $\downarrow$ & -152.04M & -777.73K & 90.7\% & 91.2\% \\
          &       &       & SR-init & 67.91  & 64.90  & -3.01 $\downarrow$ & -152.04M & -777.73K & 90.7\% & 91.2\% \\
          &       &       & Ours  & 67.91  & 64.93  & \textbf{-2.98} $\downarrow$ & -152.04M & -777.73K & 90.7\% & 91.2\% \\
\cmidrule{2-11}          & \multirow{4}[1]{*}{RML2018.01a-high} & \multirow{4}[1]{*}{89\%} & Random & 89.70  & 83.45  & -6.25 $\downarrow$ & -304.09M & -777.73K & 90.7\% & 91.1\% \\
          &       &       & LCP   & 89.70  & 82.70  & -7.00 $\downarrow$ & -304.09M & -777.73K & 90.7\% & 91.1\% \\
          &       &       & SR-init & 89.70  & 82.72  & -6.98 $\downarrow$ & -304.09M & -777.73K & 90.7\% & 91.1\% \\
          &       &       & Ours  & 89.70  & 84.07  & \textbf{-5.63} $\downarrow$ & -304.09M & -777.73K & 90.7\% & 91.1\% \\
              \midrule
    \multirow{12}[4]{*}{ResNet110} & \multirow{4}[1]{*}{RML2016.10a} & \multirow{4}[1]{*}{93\%} & Random & 61.97  & 60.55  & -1.42 $\downarrow$ & -78.41M & -1.58M & 92.6\% & 91.3\% \\
          &       &       & LCP   & 61.97  & 61.04  & \textbf{-0.93} $\downarrow$ & -78.41M & -1.58M & 92.6\% & 91.3\% \\
          &       &       & SR-init & 61.97  & 60.82  & -1.15 $\downarrow$ & -78.41M & -1.58M & 92.6\% & 91.3\% \\
          &       &       & Ours  & 61.97  & 61.00  & -0.97 $\downarrow$ & -78.41M & -1.58M & 92.6\% & 91.3\% \\
\cmidrule{2-11}          & \multirow{4}[2]{*}{Sig2019-12} & \multirow{4}[2]{*}{93\%} & Random & 65.67  & 65.07  & -0.60 $\downarrow$ & -313.62M & -1.58M & 92.6\% & 91.3\% \\
          &       &       & LCP   & 65.67  & 65.74  & -0.07 $\downarrow$ & -317.34M & -1.64M & 94.0\% & 94.8\% \\
          &       &       & SR-init & 65.67  & 65.49  & -0.18 $\downarrow$ & -313.62M & -1.58M & 92.6\% & 91.3\% \\
          &       &       & Ours  & 65.67  & 65.90  & \textbf{0.23} $\uparrow$ & -313.62M & -1.58M & 92.6\% & 91.3\% \\
\cmidrule{2-11}          & \multirow{4}[1]{*}{RML2018.01a-high} & \multirow{4}[1]{*}{93\%} & Random & 86.73  & 82.68  & -4.05 $\downarrow$ & -646.19M & -1.66M & 95.4\% & 96.0\% \\
          &       &       & LCP   & 86.73  & 83.37  & -3.36 $\downarrow$ & -646.19M & -1.66M & 95.4\% & 96.0\% \\
          &       &       & SR-init & 86.73  & 82.53  & -4.20 $\downarrow$ & -646.19M & -1.66M & 95.4\% & 96.0\% \\
          &       &       & Ours  & 86.73  & 83.79  & \textbf{-2.94} $\downarrow$ & -646.19M & -1.66M & 95.4\% & 96.0\% \\
              \midrule
    \multirow{12}[5]{*}{VGG16} & \multirow{4}[1]{*}{RML2016.10a} & \multirow{4}[1]{*}{67\%} & Random & 58.70  & 57.72  & -0.98 $\downarrow$ & -0.81G & -6.08M & 90.0\% & 68.7\% \\
          &       &       & LCP   & 58.70  & 57.70  & -1.00 $\downarrow$ & -0.81G & -6.08M & 90.0\% & 68.7\% \\
          &       &       & SR-init & 58.70  & 57.83  & -0.87 $\downarrow$ & -0.81G & -6.08M & 90.0\% & 68.7\% \\
          &       &       & Ours  & 58.70  & 57.91  & \textbf{-0.79} $\downarrow$ & -0.81G & -6.08M & 90.0\% & 68.7\% \\
\cmidrule{2-11}          & \multirow{4}[2]{*}{Sig2019-12} & \multirow{4}[2]{*}{67\%} & Random & 65.32  & 63.28  & -2.04 $\downarrow$ & -2.93G & -6.09M & 89.9\% & 50.8\% \\
          &       &       & LCP   & 65.32  & 63.05  & -2.27 $\downarrow$ & -2.93G & -6.09M & 89.9\% & 50.8\% \\
          &       &       & SR-init & 65.32  & 63.28  & -2.04 $\downarrow$ & -2.93G & -6.09M & 89.9\% & 50.8\% \\
          &       &       & Ours  & 65.32  & 64.43  & \textbf{-0.89} $\downarrow$ & -2.93G & -6.09M & 89.9\% & 50.8\% \\
\cmidrule{2-11}          & \multirow{4}[2]{*}{RML2018.01a-high} & \multirow{4}[2]{*}{44\%} & Random & 83.65  & 80.69  & -2.96 $\downarrow$ & -3.71G & -5.91M & 58.1\% & 36.5\% \\
          &       &       & LCP   & 83.65  & 81.12  & -2.53 $\downarrow$ & -4.42G & -5.46M & 69.2\% & 33.7\% \\
          &       &       & SR-init & 83.65  & 81.57  & -2.08 $\downarrow$ & -3.71G & -5.91M & 58.1\% & 36.5\% \\
          &       &       & Ours  & 83.65  & 82.58  & \textbf{-1.07} $\downarrow$ & -3.71G & -5.91M & 58.1\% & 36.5\% \\
    \bottomrule
    \end{tabular}}
  \label{layerpruning}%
\end{table*}%

\textbf{Comparison to layer pruning methods.} In the previous paragraph, we demonstrate the superiority of our method compared to existing channel pruning methods. Here, we make a comprehensive comparison  with some layer pruning methods such as random layer pruning, LCP-based pruning methods~\cite{chen2018shallowing,wang2019dbp} and SR-init~\cite{tang2023sr}. Since LCP-based pruning methods utilize different knowledge distillation techniques to recover performance, our method does not. Therefore, to ensure a fair comparison, we do not use the knowledge distillation strategy for LCP-based pruning methods. Similarly, we use the highest pruning rate achievable by our method for the experiments. Except for ResNet56 on RML2016.10a, which uses $k=4$, all other models use $k=3$ by default. The pruning rates of FLOPs and parameters are nearly the same under the same pruning rate. Therefore, accuracy can be a real reflection of the effectiveness of layer pruning methods. As shown in \cref{layerpruning}, our method outperforms all layer pruning baselines on Sig2019-12 and RML2018.01a-high. When pruning ResNet110 on RML2016.10a, our method achieves negligible lower accuracy than SR-init (-0.93\% vs. -0.97\%). While our method achieves the least performance degradation when pruning on the other two datasets (0.23\% vs. -0.18\% for SR-init on Sig2019-12, -2.94\% vs. -4.20\% for SR-init on RML2018.01a-high). In general, these experimental results suggest the effectiveness of our method.

\subsection{Additional Experiments and Analyses}
\textbf{Comparing with brute force search.} In this paper, we first divide the pre-trained model into multiple blocks based on representation similarity, and then search for the optimal layer combination block by block. A naive way is to use brute force search to obtain the optimal combination. Therefore, we compare our strategy with brute force search. For ease of understanding, let's take a $20$-layer model as an example. The search space of brute force search is $\sum_{i=1}^{20} C_{20}^i=2^{20} - 1=1,048,575$. However, suppose this $20$-layer model is divided into three blocks with $5$, $7$, and $8$ layers respectively, the search space is drastically reduced to $\sum_{i=1}^{5}+\sum_{i=1}^{7}+\sum_{i=1}^{8}=413$, which demonstrates the necessity of model partition.

\begin{figure*}[t]
	\centering
    \centering
    \includegraphics[width=0.99\textwidth]{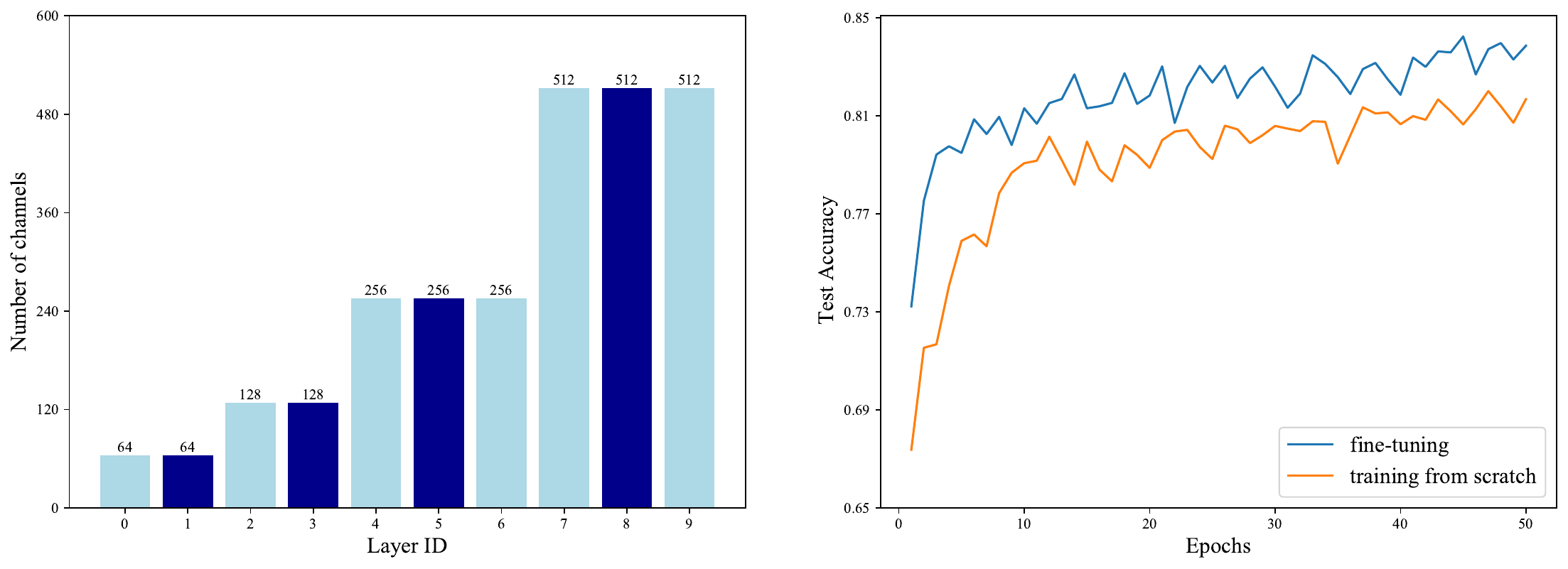} 
 \caption{Additional experiments. Left: visualizations of selected layers of VGG16 on RML2016.10a-high. Right: test accuracy curves of ResNet56 on RML2018.01a-high for training from scratch and fine-tuning.}
 \label{fig:vis}
 % \vspace{-2mm}
\end{figure*}

\textbf{Visualizations of selected layers.} Here, we show the pruned model (VGG16 on RML2016.10a-high) obtained using our layer pruning method. As shown on the left side of \cref{fig:vis}, the bars with transparent color represent the layers considered to have less contribution and need to be pruned.

\textbf{Fine-tuning vs. training from scratch.} In this paper, we fine-tune the compact model by retaining some parameters from the original pre-trained model, instead of randomly initializing all parameters (i.e., training from scratch). Therefore, in order to verify the superiority of fine-tuning, we conduct experiments on these two methods respectively. Specifically, we plot the test accuracy curves of the two methods using ResNet56 on RML2018.01a-high, as shown in the right side of \cref{fig:vis}. We observe that compared to training from scratch, fine-tuning can achieve better performance, which justifies the use of fine-tuning in our method.

% Herein, we evaluate the time cost of using partition or not. \textcolor{blue}{The number of possible layer combinations within a block increases exponentially with the number of layers in the block. For example, a block containing $5$ layers would result in a set with $31$ ($2^5 - 1$) elements. As the number of layers increases to $20$, the number of combinations increases dramatically to . It is not practical to verify the contribution of such a large number of combinations.}

\subsection{Ablation Study}
\label{sec: Ablation Study}
In this subsection, we present detailed ablation experiments on the number of blocks $k$ and the similarity metrics.
% In this subsection, we conduct detailed ablation experiments on number of blocks $N$,  utilizing different similarity metrics to replace CKA.
% To study the influence of each stage in our solution, we conduct ablation study by replacing the . We conducted detailed ablation experiments on our layer pruning methods, including functional block partitioning K-value and CKA. In order to verify the impact of the functional block partitioning K-value and CKA methods on the pruning rate and results, the experiments were conducted with other training parameters as well as the dataset and the model remaining unchanged.

% \textbf{$\operatorname{HSIC_{0}}$ v.s. $\operatorname{HSIC_{1}}$.}

\textbf{Ablation experiments on the similarity metrics.} In this paper, we uniformly use CKA to calculate the similarity between layers. However, CKA is not the only available similarity metric. To demonstrate that our method supports multiple similarity metrics, we use cosine similarity instead. Specifically, we conduct experiments on RML2016.10a-high using ResNet56. The results are presented in \cref{sim ablation}. Although different similarity metrics are used, the two methods retain the same layers due to the high pruning rate of the model, and they ultimately achieve similar accuracy. This experiment shows that our method is a general pipeline supporting multiple similarity metrics.

\textbf{Ablation experiments on number of blocks $k$.} In this paper, $k$ is a tunable parameter. To verify the impact of $k$, we conduct experiments with different $k \in \{3, 4, 5, 6, 7\}$ using ResNet110 on RML2016.10a. As shown in \cref{k ablation}, changing the value of $k$ will lead to corresponding changes in the retained layers, which slightly affects the actual pruning rate of FLOPs and parameters, as well as the performance of the pruned model. Overall, our method is robust to $k$.

% The pruning rate is taken as the maximum value under the current K-value and the results are shown in the table. Since we default to keep a minimum of one layer per function block, the more function blocks are divided, then the more layers will eventually be kept, and the corresponding pruning rate will be lower. Since the small ResNet structure contains three large layers, each of which contains the same number of BasicBlock layers, we set the minimum value of K to 3 to ensure the integrity of the ResNet structure. On ResNet110 with more layers, the relationship between the value of K and the pruning rate is still inversely proportional, but the pruning rate is the same for K of 4 and 5. This is because when only one layer is retained in each functional block of the division, these retained layers may not necessarily be present in all three major layers of ResNet, and in order not to destroy the structure of ResNet, it is only possible to increase the number of retained layers in some of the functional blocks so that there are retained layers present in all three major layers of ResNet. This leads to a situation where the pruning rate is the same for some different K values. This shows that the parameter K affects the maximum pruning rate, and we need to set reasonable parameter K according to different model structures, which can maximize the pruning rate.

% Table generated by Excel2LaTeX from sheet 'Sheet1'
\begin{table*}[t]
  \centering
  \caption{Ablation study on the similarity metrics used. PR denotes the pruning rate.}
    \resizebox{0.99\textwidth}{!}{
    \begin{tabular}{c|ccccccccc}
    \toprule
    \multicolumn{1}{c}{Model} & Original Acc(\%) & Method & PR & Acc(\%) & $\triangle$Acc(\%) & $\triangle$FLOPs & $\triangle$Params & FLOPs PR & Params PR \\
    \midrule
    \multirow{2}[2]{*}{ResNet56} & \multirow{2}[2]{*}{90.32} & CKA   & 89\%  & 90.25  & -0.07  & -38.01M & -777.72K & 90.7\% & 91.2\% \\
          &       & Cosine similarity & 89\%  & 90.35  & 0.03  & -38.01M & -777.72K & 90.7\% & 91.2\% \\
    \bottomrule
    \end{tabular}%
  \label{sim ablation}}
\end{table*}%

% Table generated by Excel2LaTeX from sheet 'Sheet1' k的消融实验，剪枝率取当前K下的最大值，数据集全信噪比的RML2016.10a									
\begin{table*}[t]
  \centering
  \caption{Ablation study on the number of blocks $k$. PR denotes the pruning rate.}
    \resizebox{0.99\textwidth}{!}{
    \begin{tabular}{c|ccccccccc}
    \toprule
    \multicolumn{1}{c}{Model} & Original Acc(\%) & k     & PR & Acc(\%) & $\triangle$Acc(\%) & $\triangle$Flops & $\triangle$Params & Flops PR & Params PR \\
    \midrule
    % \multirow{2}[2]{*}{ResNet56} & \multirow{2}[2]{*}{62.06} & 3     & 86\%  & 60.90  & -1.16  & -36.82M & -759.16K & 87.8\% & 89.0\% \\
    %       &       & 5     & 75\%  & 61.62  & -0.44  & -33.26M & -666.62K & 79.3\% & 78.2\% \\
    % \midrule
    \multirow{5}[2]{*}{ResNet110} & \multirow{5}[2]{*}{61.97} & 3     & 93\%  & 61.00  & -0.97  & -78.41M & -1.58M & 92.6\% & 91.3\% \\
          &       & 4     & 91\%  & 61.45  & -0.52  & -78.39M & -1.63M & 92.6\% & 94.2\% \\
          &       & 5     & 91\%  & 61.36  & -0.61  & -77.21M & -1.58M & 91.2\% & 91.3\% \\
          &       & 6     & 89\%  & 61.40  & -0.57  & -76.02M & -1.58M & 89.8\% & 91.3\% \\
          &       & 7     & 87\%  & 61.73  & -0.24  & -74.83M & -1.56M & 88.4\% & 90.2\% \\
    \bottomrule
    \end{tabular}}
  \label{k ablation}%
\end{table*}%

\section{Conclusion}
\label{sec:Conclusion}
In this paper, we propose a novel layer pruning method. Specifically, we first calculate the similarity matrix between layers, and then divide the pre-trained model into blocks. Subsequently, we identify layers that need to be preserved within each block based on their contribution. Finally, we reassemble the pruned blocks and fine-tune the compact model. Extensive experiments RML2016.10a, RML2016.10a-high, Sig2019-12, Sig2019-12-high and RML2018.01a demonstrate the efficiency and effectiveness of our method over a variety of state-of-the-art baselines, including channel pruning methods (RFP, FPGM, L1-norm, SFP and BNP) as well as layer pruning methods (random layer pruning, LCP-based pruning methods and SR-init).

\section*{Acknowledgments}
% This work is supported in part by the Key R\&D Program of Zhejiang under Grant 2022C01018 and Grant 61973273.
This work was partially supported by the Key R\&D Program of Zhejiang under Grant 2022C01018 and by the National Natural Science Foundation of China under Grant U21B2001.

%{\appendices
%\section*{Proof of the First Zonklar Equation}
%Appendix one text goes here.
% You can choose not to have a title for an appendix if you want by leaving the argument blank
%\section*{Proof of the Second Zonklar Equation}
%Appendix two text goes here.}
 
 % argument is your BibTeX string definitions and bibliography database(s)
%\bibliography{IEEEabrv,../bib/paper}
%
% \clearpage
\bibliographystyle{IEEEtran}
\bibliography{reference}

% Generated by IEEEtran.bst, version: 1.14 (2015/08/26)
\begin{thebibliography}{10}
\providecommand{\url}[1]{#1}
\csname url@samestyle\endcsname
\providecommand{\newblock}{\relax}
\providecommand{\bibinfo}[2]{#2}
\providecommand{\BIBentrySTDinterwordspacing}{\spaceskip=0pt\relax}
\providecommand{\BIBentryALTinterwordstretchfactor}{4}
\providecommand{\BIBentryALTinterwordspacing}{\spaceskip=\fontdimen2\font plus
\BIBentryALTinterwordstretchfactor\fontdimen3\font minus \fontdimen4\font\relax}
\providecommand{\BIBforeignlanguage}[2]{{%
\expandafter\ifx\csname l@#1\endcsname\relax
\typeout{** WARNING: IEEEtran.bst: No hyphenation pattern has been}%
\typeout{** loaded for the language `#1'. Using the pattern for}%
\typeout{** the default language instead.}%
\else
\language=\csname l@#1\endcsname
\fi
#2}}
\providecommand{\BIBdecl}{\relax}
\BIBdecl

\bibitem{walenczykowska2016type}
M.~Walenczykowska and A.~Kawalec, ``Type of modulation identification using wavelet transform and neural network,'' \emph{Bulletin of the Polish Academy of Sciences. Technical Sciences}, vol.~64, no.~1, pp. 257--261, 2016.

\bibitem{li2019wavelet}
W.~Li, Z.~Dou, L.~Qi, and C.~Shi, ``Wavelet transform based modulation classification for 5g and uav communication in multipath fading channel,'' \emph{Physical Communication}, vol.~34, pp. 272--282, 2019.

\bibitem{triantafyllakis2017phasma}
K.~Triantafyllakis, M.~Surligas, G.~Vardakis, and S.~Papadakis, ``Phasma: An automatic modulation classification system based on random forest,'' in \emph{2017 IEEE International Symposium on Dynamic Spectrum Access Networks (DySPAN)}.\hskip 1em plus 0.5em minus 0.4em\relax IEEE, 2017, pp. 1--3.

\bibitem{vuvcic2017cyclic}
D.~Vu{\v{c}}i{\'c}, S.~Vukoti{\'c}, and M.~Eri{\'c}, ``Cyclic spectral analysis of ofdm/oqam signals,'' \emph{AEU-International Journal of Electronics and Communications}, vol.~73, pp. 139--143, 2017.

\bibitem{abdelmutalab2016automatic}
A.~Abdelmutalab, K.~Assaleh, and M.~El-Tarhuni, ``Automatic modulation classification based on high order cumulants and hierarchical polynomial classifiers,'' \emph{Physical Communication}, vol.~21, pp. 10--18, 2016.

\bibitem{o2018over}
T.~J. O’Shea, T.~Roy, and T.~C. Clancy, ``Over-the-air deep learning based radio signal classification,'' \emph{IEEE Journal of Selected Topics in Signal Processing}, vol.~12, no.~1, pp. 168--179, 2018.

\bibitem{simonyan2014very}
K.~Simonyan and A.~Zisserman, ``Very deep convolutional networks for large-scale image recognition,'' \emph{arXiv preprint arXiv:1409.1556}, 2014.

\bibitem{o2016convolutional}
T.~J. O’Shea, J.~Corgan, and T.~C. Clancy, ``Convolutional radio modulation recognition networks,'' in \emph{Engineering Applications of Neural Networks: 17th International Conference, EANN 2016, Aberdeen, UK, September 2-5, 2016, Proceedings 17}.\hskip 1em plus 0.5em minus 0.4em\relax Springer, 2016, pp. 213--226.

\bibitem{chen2021signet}
Z.~Chen, H.~Cui, J.~Xiang, K.~Qiu, L.~Huang, S.~Zheng, S.~Chen, Q.~Xuan, and X.~Yang, ``Signet: A novel deep learning framework for radio signal classification,'' \emph{IEEE Transactions on Cognitive Communications and Networking}, vol.~8, no.~2, pp. 529--541, 2021.

\bibitem{zhang2018systematic}
T.~Zhang, S.~Ye, K.~Zhang, J.~Tang, W.~Wen, M.~Fardad, and Y.~Wang, ``A systematic dnn weight pruning framework using alternating direction method of multipliers,'' in \emph{ECCV}, 2018, pp. 184--199.

\bibitem{ma2020pconv}
X.~Ma, F.-M. Guo, W.~Niu, X.~Lin, J.~Tang, K.~Ma, B.~Ren, and Y.~Wang, ``Pconv: The missing but desirable sparsity in dnn weight pruning for real-time execution on mobile devices,'' in \emph{AAAI}, vol.~34, no.~04, 2020, pp. 5117--5124.

\bibitem{meng2020pruning}
F.~Meng, H.~Cheng, K.~Li, H.~Luo, X.~Guo, G.~Lu, and X.~Sun, ``Pruning filter in filter,'' \emph{NeurIPS}, vol.~33, pp. 17\,629--17\,640, 2020.

\bibitem{chen2023rgp}
Z.~Chen, J.~Xiang, Y.~Lu, Q.~Xuan, Z.~Wang, G.~Chen, and X.~Yang, ``Rgp: Neural network pruning through regular graph with edges swapping,'' \emph{IEEE Transactions on Neural Networks and Learning Systems}, 2023.

\bibitem{lu2022understanding}
Y.~Lu, W.~Yang, Y.~Zhang, Z.~Chen, J.~Chen, Q.~Xuan, Z.~Wang, and X.~Yang, ``Understanding the dynamics of dnns using graph modularity,'' in \emph{European Conference on Computer Vision}.\hskip 1em plus 0.5em minus 0.4em\relax Springer, 2022, pp. 225--242.

\bibitem{tang2023sr}
H.~Tang, Y.~Lu, and Q.~Xuan, ``Sr-init: An interpretable layer pruning method,'' in \emph{ICASSP}.\hskip 1em plus 0.5em minus 0.4em\relax IEEE, 2023, pp. 1--5.

\bibitem{tanaka2020pruning}
H.~Tanaka, D.~Kunin, D.~L. Yamins, and S.~Ganguli, ``Pruning neural networks without any data by iteratively conserving synaptic flow,'' \emph{Advances in neural information processing systems}, vol.~33, pp. 6377--6389, 2020.

\bibitem{o2016radio}
T.~J. O'shea and N.~West, ``Radio machine learning dataset generation with gnu radio,'' in \emph{Proceedings of the GNU Radio Conference}, vol.~1, no.~1, 2016.

\bibitem{shao2021filter}
L.~Shao, H.~Zuo, J.~Zhang, Z.~Xu, J.~Yao, Z.~Wang, and H.~Li, ``Filter pruning via measuring feature map information,'' \emph{Sensors}, vol.~21, no.~19, p. 6601, 2021.

\bibitem{he2019filter}
Y.~He, P.~Liu, Z.~Wang, Z.~Hu, and Y.~Yang, ``Filter pruning via geometric median for deep convolutional neural networks acceleration,'' in \emph{Proceedings of the IEEE/CVF conference on computer vision and pattern recognition}, 2019, pp. 4340--4349.

\bibitem{li2016pruning}
H.~Li, A.~Kadav, I.~Durdanovic, H.~Samet, and H.~P. Graf, ``Pruning filters for efficient convnets,'' \emph{arXiv preprint arXiv:1608.08710}, 2016.

\bibitem{he2018soft}
Y.~He, G.~Kang, X.~Dong, Y.~Fu, and Y.~Yang, ``Soft filter pruning for accelerating deep convolutional neural networks,'' in \emph{Proceedings of the 27th International Joint Conference on Artificial Intelligence}, 2018, pp. 2234--2240.

\bibitem{chen2023channel}
Z.~Chen, Z.~Wang, X.~Gao, J.~Zhou, D.~Xu, S.~Zheng, Q.~Xuan, and X.~Yang, ``Channel pruning method for signal modulation recognition deep learning models,'' \emph{IEEE Transactions on Cognitive Communications and Networking}, 2023.

\bibitem{chen2018shallowing}
S.~Chen and Q.~Zhao, ``Shallowing deep networks: Layer-wise pruning based on feature representations,'' \emph{TPAMI}, vol.~41, no.~12, pp. 3048--3056, 2018.

\bibitem{wang2019dbp}
W.~Wang, S.~Zhao, M.~Chen, J.~Hu, D.~Cai, and H.~Liu, ``Dbp: Discrimination based block-level pruning for deep model acceleration,'' \emph{arXiv preprint arXiv:1912.10178}, 2019.

\bibitem{hardoon2004canonical}
D.~R. Hardoon, S.~Szedmak, and J.~Shawe-Taylor, ``Canonical correlation analysis: An overview with application to learning methods,'' \emph{Neural computation}, vol.~16, no.~12, pp. 2639--2664, 2004.

\bibitem{raghu2017svcca}
M.~Raghu, J.~Gilmer, J.~Yosinski, and J.~Sohl-Dickstein, ``Svcca: Singular vector canonical correlation analysis for deep learning dynamics and interpretability,'' \emph{Advances in neural information processing systems}, vol.~30, 2017.

\bibitem{morcos2018insights}
A.~Morcos, M.~Raghu, and S.~Bengio, ``Insights on representational similarity in neural networks with canonical correlation,'' \emph{Advances in neural information processing systems}, vol.~31, 2018.

\bibitem{kornblith2019similarity}
S.~Kornblith, M.~Norouzi, H.~Lee, and G.~Hinton, ``Similarity of neural network representations revisited,'' in \emph{International conference on machine learning}.\hskip 1em plus 0.5em minus 0.4em\relax PMLR, 2019, pp. 3519--3529.

\bibitem{chen2021graph}
Z.~Chen, Y.~Lu, J.~Hu, W.~Yang, Q.~Xuan, Z.~Wang, and X.~Yang, ``Graph-based similarity of neural network representations,'' \emph{arXiv preprint arXiv:2111.11165}, 2021.

\bibitem{aghli2021combining}
N.~Aghli and E.~Ribeiro, ``Combining weight pruning and knowledge distillation for cnn compression,'' in \emph{CVPR}, 2021, pp. 3191--3198.

\bibitem{molchanov2019importance}
P.~Molchanov, A.~Mallya, S.~Tyree, I.~Frosio, and J.~Kautz, ``Importance estimation for neural network pruning,'' in \emph{CVPR}, 2019, pp. 11\,264--11\,272.

\bibitem{he2017channel}
Y.~He, X.~Zhang, and J.~Sun, ``Channel pruning for accelerating very deep neural networks,'' in \emph{ICCV}, 2017, pp. 1389--1397.

\bibitem{zhuang2018discrimination}
Z.~Zhuang, M.~Tan, B.~Zhuang, J.~Liu, Y.~Guo, Q.~Wu, J.~Huang, and J.~Zhu, ``Discrimination-aware channel pruning for deep neural networks,'' \emph{Advances in neural information processing systems}, vol.~31, 2018.

\bibitem{huang2018data}
Z.~Huang and N.~Wang, ``Data-driven sparse structure selection for deep neural networks,'' in \emph{ECCV}, 2018, pp. 304--320.

\bibitem{park2016faster}
J.~Park, S.~Li, W.~Wen, P.~T.~P. Tang, H.~Li, Y.~Chen, and P.~Dubey, ``Faster cnns with direct sparse convolutions and guided pruning,'' in \emph{ICLR}, 2016.

\bibitem{han2016eie}
S.~Han, X.~Liu, H.~Mao, J.~Pu, A.~Pedram, M.~A. Horowitz, and W.~J. Dally, ``Eie: Efficient inference engine on compressed deep neural network,'' \emph{ACM SIGARCH Computer Architecture News}, vol.~44, no.~3, pp. 243--254, 2016.

\bibitem{elkerdawy2020filter}
S.~Elkerdawy, M.~Elhoushi, A.~Singh, H.~Zhang, and N.~Ray, ``To filter prune, or to layer prune, that is the question,'' in \emph{Proceedings of the Asian Conference on Computer Vision}, 2020.

\bibitem{deng2009imagenet}
J.~Deng, W.~Dong, R.~Socher, L.-J. Li, K.~Li, and L.~Fei-Fei, ``Imagenet: A large-scale hierarchical image database,'' in \emph{2009 IEEE conference on computer vision and pattern recognition}.\hskip 1em plus 0.5em minus 0.4em\relax Ieee, 2009, pp. 248--255.

\bibitem{elkerdawy2020one}
S.~Elkerdawy, M.~Elhoushi, A.~Singh, H.~Zhang, and N.~Ray, ``One-shot layer-wise accuracy approximation for layer pruning,'' in \emph{2020 IEEE International Conference on Image Processing (ICIP)}.\hskip 1em plus 0.5em minus 0.4em\relax IEEE, 2020, pp. 2940--2944.

\bibitem{dror2021layer}
A.~B. Dror, N.~Zehngut, A.~Raviv, E.~Artyomov, R.~Vitek, and R.~Jevnisek, ``Layer folding: Neural network depth reduction using activation linearization,'' \emph{arXiv preprint arXiv:2106.09309}, 2021.

\bibitem{wu2023efficient}
J.~Wu, D.~Zhu, L.~Fang, Y.~Deng, and Z.~Zhong, ``Efficient layer compression without pruning,'' \emph{IEEE Transactions on Image Processing}, 2023.

\bibitem{gretton2005measuring}
A.~Gretton, O.~Bousquet, A.~Smola, and B.~Sch{\"o}lkopf, ``Measuring statistical dependence with hilbert-schmidt norms,'' in \emph{International conference on algorithmic learning theory}.\hskip 1em plus 0.5em minus 0.4em\relax Springer, 2005, pp. 63--77.

\bibitem{song2012feature}
L.~Song, A.~Smola, A.~Gretton, J.~Bedo, and K.~Borgwardt, ``Feature selection via dependence maximization.'' \emph{Journal of Machine Learning Research}, vol.~13, no.~5, 2012.

\bibitem{he2015delving}
K.~He, X.~Zhang, S.~Ren, and J.~Sun, ``Delving deep into rectifiers: Surpassing human-level performance on imagenet classification,'' in \emph{Proceedings of the IEEE international conference on computer vision}, 2015, pp. 1026--1034.

\bibitem{blossom2004gnu}
E.~Blossom, ``Gnu radio: tools for exploring the radio frequency spectrum,'' \emph{Linux journal}, vol. 2004, no. 122, p.~4, 2004.

\bibitem{he2016deep}
K.~He, X.~Zhang, S.~Ren, and J.~Sun, ``Deep residual learning for image recognition,'' in \emph{Proceedings of the IEEE conference on computer vision and pattern recognition}, 2016, pp. 770--778.

\bibitem{fletcher2008robust}
P.~T. Fletcher, S.~Venkatasubramanian, and S.~Joshi, ``Robust statistics on riemannian manifolds via the geometric median,'' in \emph{2008 IEEE Conference on Computer Vision and Pattern Recognition}.\hskip 1em plus 0.5em minus 0.4em\relax IEEE, 2008, pp. 1--8.

\end{thebibliography}
% \vspace{-5mm}

\begin{IEEEbiography}[{\includegraphics[width=1in,height=1.25in,clip,keepaspectratio]{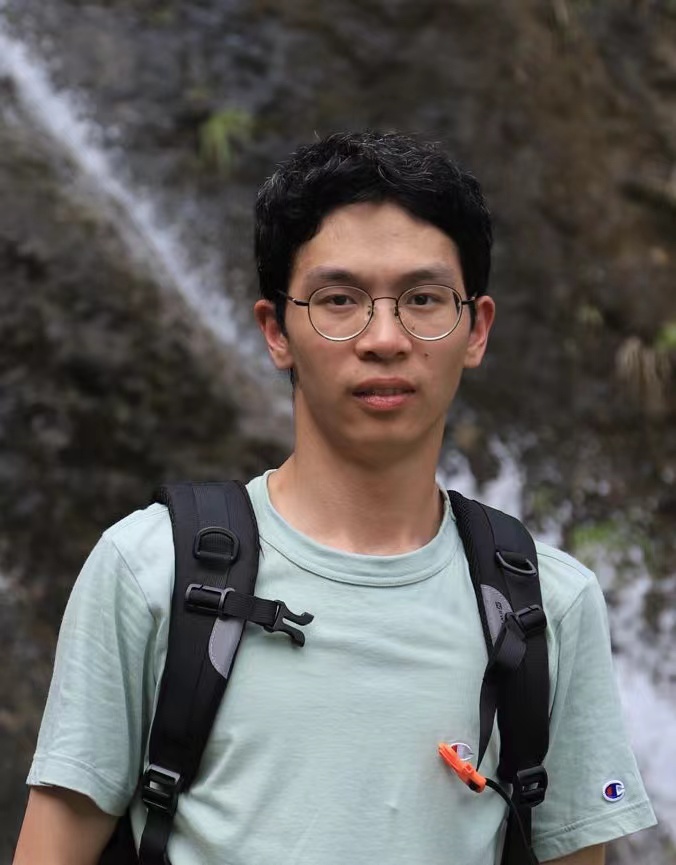}}]{Yao Lu}
received his B.S. degree from Zhejiang University of Technology and is currently pursuing a Ph.D. in control science and engineering at Zhejiang University of Technology. He has published several academic papers in international conferences and journals, including ECCV and TNNLS. His research interests include deep learning and computer vision, with a focus on explainable artificial intelligence and model compression.
\end{IEEEbiography} 
\vspace{-8mm}

\begin{IEEEbiography}[{\includegraphics[width=1in,height=1.25in,clip,keepaspectratio]{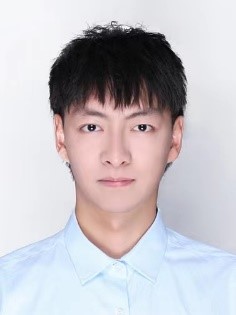}}]{Yutao Zhu}
received his bachelor's degree from the School of Electrical and Electronic Engineering at Wenzhou University in 2023. He is currently a master's student at the School of Information Engineering at Zhejiang University of Technology. His research interests include signal processing and lightweight neural networks.
\end{IEEEbiography} 

\vspace{-8mm}
\begin{IEEEbiography}[{\includegraphics[width=1in,height=1.25in,clip,keepaspectratio]{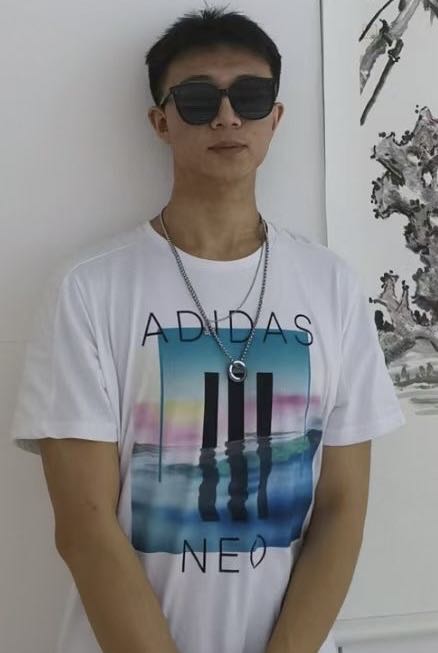}}]{Yuqi Li}
received his bachelor's degree from Southwest University and is currently engaged as a research intern at the Institute of Computing Technology, Chinese Academy of Sciences. His research interests include model compression and medical imaging.
\end{IEEEbiography} 

\begin{IEEEbiography}[{\includegraphics[width=1in,height=1.25in,clip,keepaspectratio]{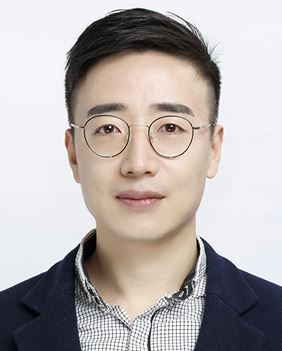}}]{Dongwei Xu}
(Member, IEEE) received the B.E. and Ph.D. degrees from the State Key Laboratory of Rail Traffic Control and Safety, Beijing Jiaotong University, Beijing, China, in 2008 and 2014, respectively. He is currently an Associate Professor with the Institute of Cyberspace Security, Zhejiang University of Technology, Hangzhou, China. His research interests include intelligent transportation Control, management, and traffic safety engineering.
\end{IEEEbiography}

\begin{IEEEbiography}[{\includegraphics[width=1in,height=1.25in,clip,keepaspectratio]{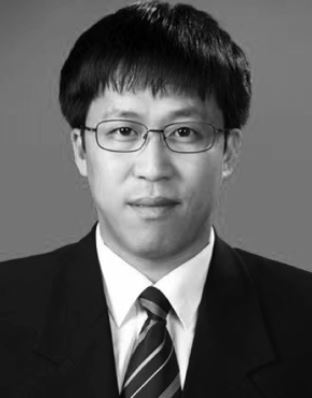}}]{Yun Lin}
(Member, IEEE) received the B.S. degree in electrical engineering from Dalian Maritime University, Dalian, China, in 2003, the M.S. degree in communication and information system from the Harbin Institute of Technology, Harbin, China, in 2005, and the Ph.D. degree in communication and information system from Harbin Engineering University, Harbin, in 2010. From 2014 to 2015, he was a Research Scholar with Wright State University, Dayton, OH, USA. He is currently a Full Professor with the College of Information and Communication Engineering, Harbin Engineering University. He has authored or coauthored more than 200 international peer-reviewed journal/conference papers, such as IEEE Transactions on Industrial Informatics, IEEE Transactions on Communications, IEEE Internet of Things Journal, IEEE Transactions on Vehicular Technology, IEEE Transactions on Cognitive Communications and Networking, TR, INFOCOM, GLOBECOM, ICC, VTC, and ICNC. His current research interests include machine learning and data analytics over wireless networks, signal processing and analysis, cognitive radio and software-defined radio, artificial intelligence, and pattern recognition.
\end{IEEEbiography}

\begin{IEEEbiography}[{\includegraphics[width=1in,height=1.25in,clip,keepaspectratio]{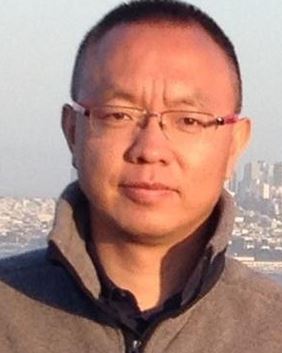}}]{Qi Xuan}
(Senior Member, IEEE) received the B.S. and Ph.D. degrees in control theory and engineering from Zhejiang University, Hangzhou, China, in 2003 and 2008, respectively. He was a Postdoctoral Researcher with the Department of Information Science and Electronic Engineering, Zhejiang University from 2008 to 2010, and a Research Assistant with the Department of Electronic Engineering, City University of Hong Kong, Hong Kong, in 2010 and 2017, respectively. From 2012 to 2014, he was a Postdoctoral Fellow with the Department of Computer Science, University of California at Davis, Davis, CA, USA. He is currently a Professor with the Institute of Cyberspace Security, College of Information Engineering, Zhejiang University of Technology, Hangzhou, and also with the PCL Research Center of Networks and Communications, Peng Cheng Laboratory, Shenzhen, China. He is also with Utron Technology Company Ltd., Xi’an, China, as a Hangzhou Qianjiang Distinguished Expert. His current research interests include network science, graph data mining, cyberspace security, machine learning, and computer vision.
\end{IEEEbiography}

\begin{IEEEbiography}[{\includegraphics[width=1in,height=1.25in,clip,keepaspectratio]{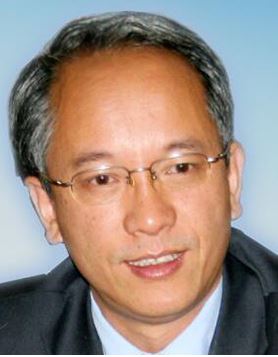}}]{Xiaoniu Yang}
is currently a Chief Scientist with the Science and Technology on Communication Information Security Control Laboratory, Jiaxing, China. He published the first software radio book in China [X. Yang, C. Lou, and J. Xu, Software Radio Principles and Applications, Publishing House of Electronics Industry, 2001 (in Chinese)]. His current research interests are software-defined satellite, big data for radio signals, and deep-learning-based signal processing. He is also an Academician of the Chinese Academy of Engineering and a Fellow of the Chinese Institute of Electronics.
\end{IEEEbiography}

\vfill

\end{document}